\documentclass{article}

\usepackage{preprint}

\usepackage[utf8]{inputenc}
\usepackage[T1]{fontenc}
\usepackage{hyperref}
\usepackage{url}
\usepackage{booktabs}
\usepackage{amsfonts}
\usepackage{nicefrac}
\usepackage{microtype}
\usepackage{xcolor}

\usepackage{graphicx}
\usepackage{subcaption}
\usepackage{multirow}

\usepackage{amsmath}
\usepackage{amssymb}
\usepackage{mathtools}
\usepackage{amsthm}
\usepackage{wrapfig}

\usepackage[capitalize,noabbrev]{cleveref}

\usepackage{amsmath,amsfonts,bm}

\DeclareMathOperator{\defeq}{\stackrel{\text{def}}{\;=\;}}

\def\eqref#1{equation~\ref{#1}}

\def\1{\bm{1}}

\DeclareMathAlphabet{\mathsfit}{\encodingdefault}{\sfdefault}{m}{sl}
\SetMathAlphabet{\mathsfit}{bold}{\encodingdefault}{\sfdefault}{bx}{n}

\theoremstyle{plain}

\theoremstyle{definition}

\theoremstyle{remark}

\definecolor{legendblue}{RGB}{31, 119, 180}
\definecolor{legendgreen}{RGB}{44, 160, 44}

\usepackage[disable,textsize=tiny]{todonotes}

\title{Test Set Contamination of Generative Evaluations}

\author{%
  \parbox{\dimexpr\textwidth-2\tabcolsep\relax}{\centering
    Rylan Schaeffer\textsuperscript{1}\thanks{Corresponding author: \texttt{rschaef@cs.stanford.edu}.}\quad
    Joshua Kazdan\textsuperscript{2}\quad
    Baber Abbasi\textsuperscript{3}\\[3pt]
    Ken Ziyu Liu\textsuperscript{1}\quad
    Brando Miranda\textsuperscript{1}\quad
    Ahmed Ahmed\textsuperscript{1}\\[3pt]
    Fazl Barez\textsuperscript{4,5}\quad
    Yegor Denisov-Blanch\textsuperscript{1}\quad
    Abhay Puri\textsuperscript{6}\\[3pt]
    Stella Biderman\textsuperscript{3}\quad
    Niloofar Mireshghallah\textsuperscript{7}\quad
    Sanmi Koyejo\textsuperscript{1}\thanks{Corresponding author: \texttt{sanmi@cs.stanford.edu}.}\\[7pt]
    {\normalfont\small
      \textsuperscript{1}Stanford Computer Science\quad
      \textsuperscript{2}Stanford Statistics\\
      \textsuperscript{3}EleutherAI\quad
      \textsuperscript{4}University of Oxford\quad
      \textsuperscript{5}Martian\\
      \textsuperscript{6}ServiceNow Research\quad
      \textsuperscript{7}Carnegie Mellon University
    }
  }
}

\begin{document}

\maketitle

\begin{abstract}
    Test set contamination -- the inclusion of benchmarks in pretraining data -- is a critical challenge to the trustworthy evaluation of frontier AI systems.
    While its impact on \emph{discriminative} evaluations is well-studied, contamination on \emph{generative} evaluations remains underexplored.
    We quantitatively assess these effects across the language model lifecycle by pretraining models (up to 344M parameters) contaminated with varying numbers of MATH test set replicas.
    While performance expectedly improves with contamination and model size, the improvements stem from superficial memorization, not generalization.
    Our scaling law analysis reveals a fundamental breach: including even a single test set replica enables models to achieve lower loss than the irreducible error of training on the uncontaminated corpus.
    We then study additional training: overtraining with fresh data dilutes contamination effects, whereas supervised finetuning on the training set improves performance for low contamination but degrades it for high contamination.
    At inference, we identify three distinct regimes of memorization -- exponential decoherence, brittle memorization, deterministic lock-in -- governed by solution length and sampling temperature.
    Finally, we identify and fix a critical implementation error in EleutherAI's LM Evaluation Harness that previously underreported mathematical reasoning performance. By characterizing how generation and memorization interact, we highlight new considerations for trustworthy AI evaluation.
\end{abstract}

\section{Introduction}
\label{sec:introduction}

As frontier AI systems are pretrained on web-scale data, test set contamination has become a critical concern for accurately assessing capabilities \citep{sainz2023nlp,schaeffer2023pretrainingtestsetneed,xu2024benchmarkdatacontaminationlarge,deng2024unveiling,reuel2025open}.
Evaluation aims to measure generalization on unseen tasks, yet the sheer scale of modern pretraining makes such contamination highly likely \citep{brown2020language, du2022glam, wei2022finetuned,chowdhery2022palmscalinglanguagemodeling,touvron2023llama2openfoundation, penedo2024finewebdatasetsdecantingweb}.

Prior research has sought to quantify the impact of test set contamination through two primary lenses. \textit{Statistical approaches} attempt to detect contamination or estimate its influence by modifying the test set, for example, by reordering, rephrasing, or replicating benchmark problems, e.g., \citep{oren2023proving, ni2025trainingbenchmarkneed, shi2024detecting, golchin2023data,golchin2024time,roberts2024to, wang2025generalization, zhang2024carefulexaminationlargelanguage}.
In comparison, \textit{controlled approaches} intentionally contaminate pretraining corpora to quantify how specific doses of leakage inflate performance, e.g., \citep{magar2022data,jiang2024investigatingdatacontaminationpretraining,oren2023proving,yao2024data, duan2024membership,wang2025generalization,kocyigit2025overestimation,bordt2025howmuch, wei2025hubblemodelsuiteadvance}.
For Related Work, see Appendix~\ref{app:sec:related_work}.

\begin{figure*}[t!]
\centering
\hspace{-1em}
\begin{minipage}[t]{0.47\textwidth}
    \vspace{0pt}
    \centering
    \includegraphics[width=\linewidth]{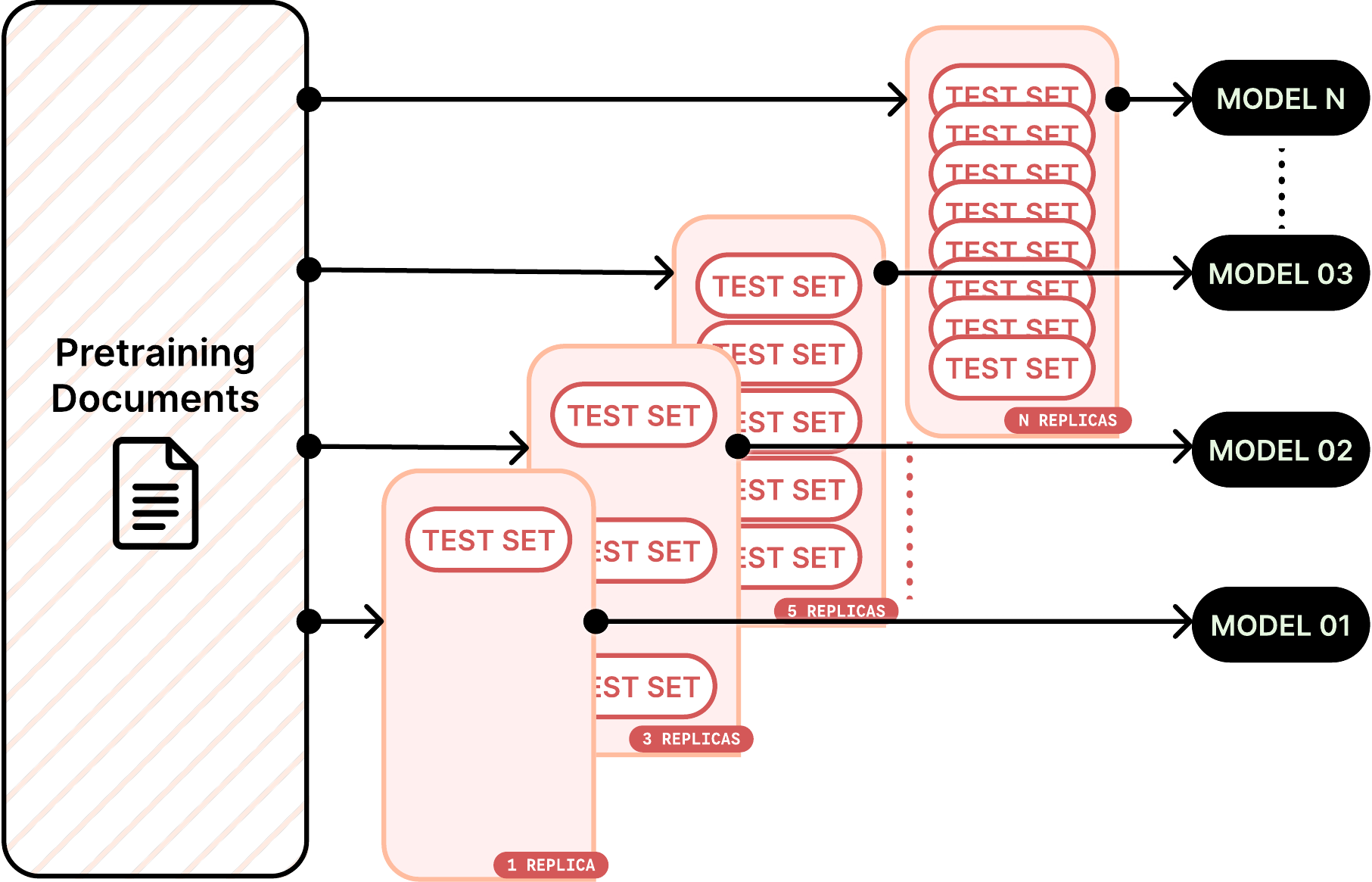}
\end{minipage}%
\hspace{0.1em}
\begin{minipage}[t]{0.49\textwidth}
    \vspace{0pt}
    \centering
    \includegraphics[width=\linewidth]{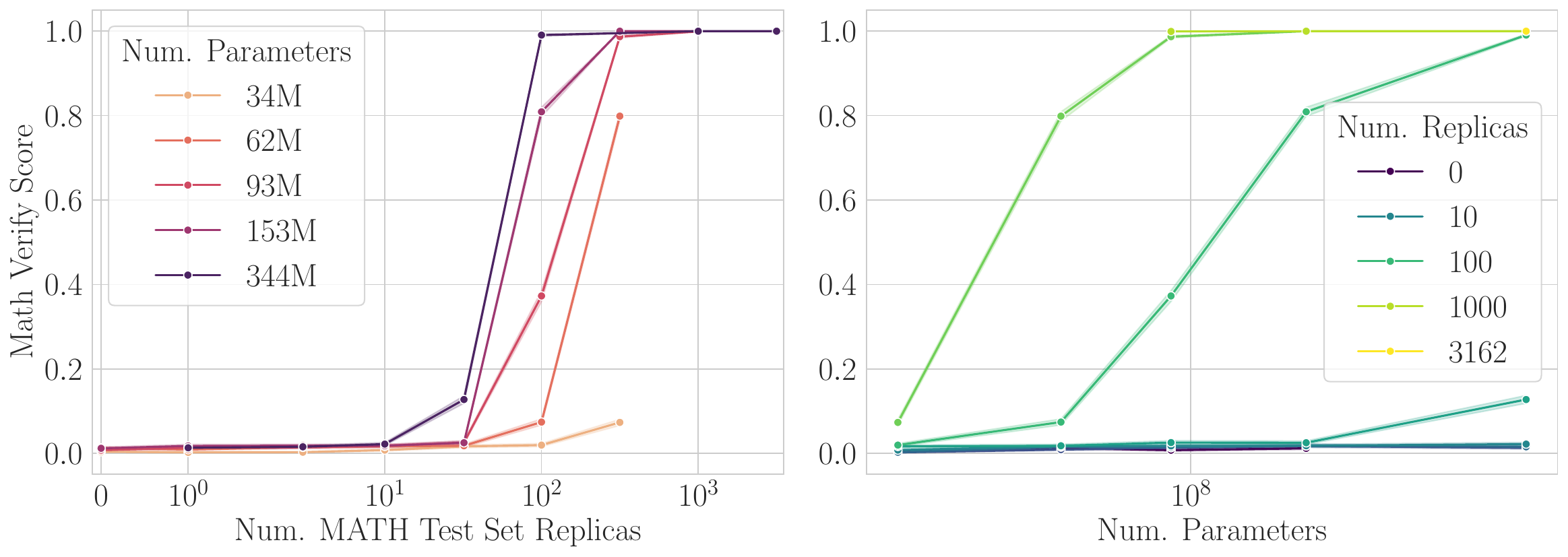}
    \includegraphics[width=\linewidth]{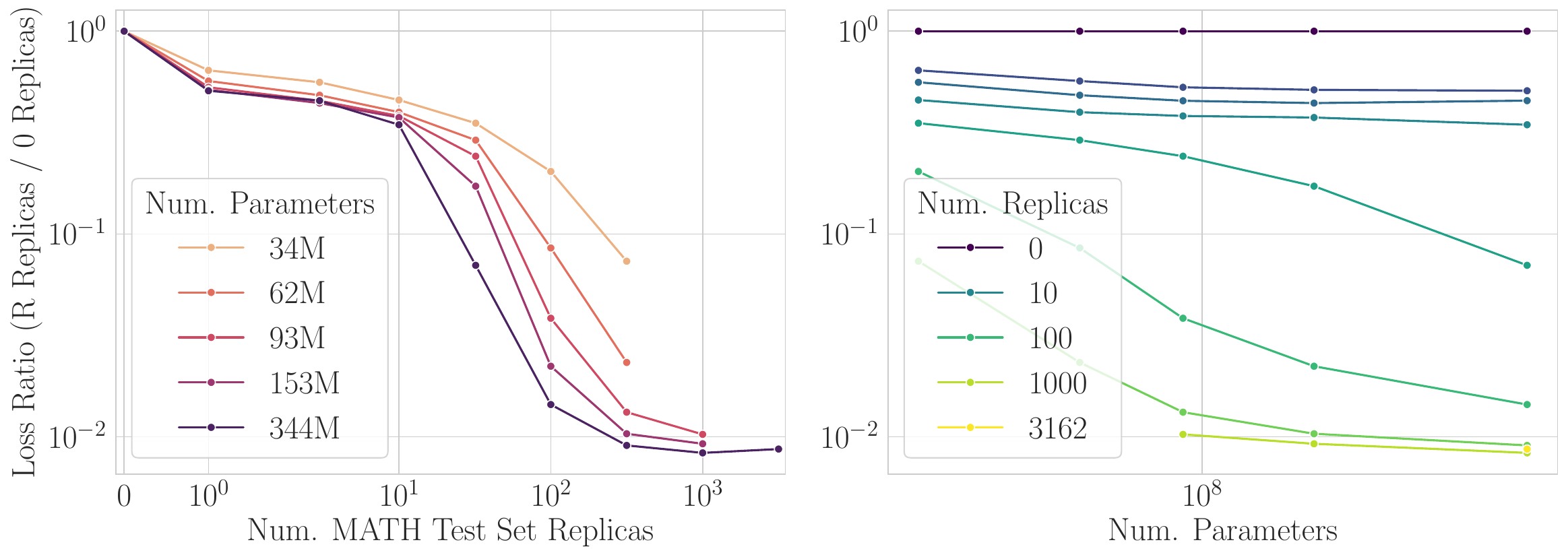}
\end{minipage}
    \caption{\textbf{Performance on Generative Benchmarks Increases with Test Set Contamination and Model Size.} We pretrained compute-optimal language models ($34$M--$344$M parameters) on corpora containing $0$--$3162$ replicas of the MATH test set, then evaluated with greedy decoding. As contamination grows, Math Verify scores rise (top) and test-set cross entropies fall (bottom), with sharp improvement around $100$ replicas. The ratio of test-set loss at $R$ replicas to loss at $0$ replicas grows with model size, so larger models benefit more from contamination at any replica count.
}
    \label{fig:math_verify_ce_temp_0}
\end{figure*}

While foundational, these investigations have predominantly focused on \textit{discriminative} benchmarks like classification or multiple-choice question-answering (MCQA). For example, \citet{magar2022data} used SST-2 \citep{socher2013sst} (classification). \citet{jiang2024investigatingdatacontaminationpretraining} used SST-2 (classification), MMLU \citep{hendrycks2021measuring} (MCQA), SQuAD \citep{rajpurkar2016squad} (MCQA), and CNN/Daily Mail (fill-in-the-middle) \citep{nallapati2016abstractive}. \citet{oren2023proving} used 7 MCQA benchmarks and 1 mathematical problem solving benchmark (GSM8K) \citep{cobbe2021trainingverifierssolvemath}, and \citet{yao2024data} used 3 MCQA benchmarks while \citet{bordt2025howmuch} used 7 MCQA benchmarks.

However, with the advancement of capabilities and the advent of reasoning models \citep{openai2024openaio1card,comanici2025gemini25pushingfrontier,xu2025largereasoningmodelssurvey}, the field is shifting towards benchmarks for which the model must generate an answer rather than choose between choices.
Whether test set contamination has the same effect on \textit{generative evaluations} is unclear.
Discriminative evaluations typically require the model to place higher probability mass on the correct choice than on a few incorrect choices \citep{gao2024evalharness, schaeffer2025elusive}, and candidate choices are often only a few tokens long.
In contrast, generative evaluations require the model to produce tens-to-thousands of tokens without straying from the memorized solution, and introduce new considerations such as the sampling temperature \citep{ackley1985learning}, the sampling algorithm (e.g., top-k \citep{fan2018topk}, top-p \citep{holtzman2020topp}) and the solution length.

\begin{figure*}[t!]
    \centering
    \includegraphics[width=0.9\linewidth]{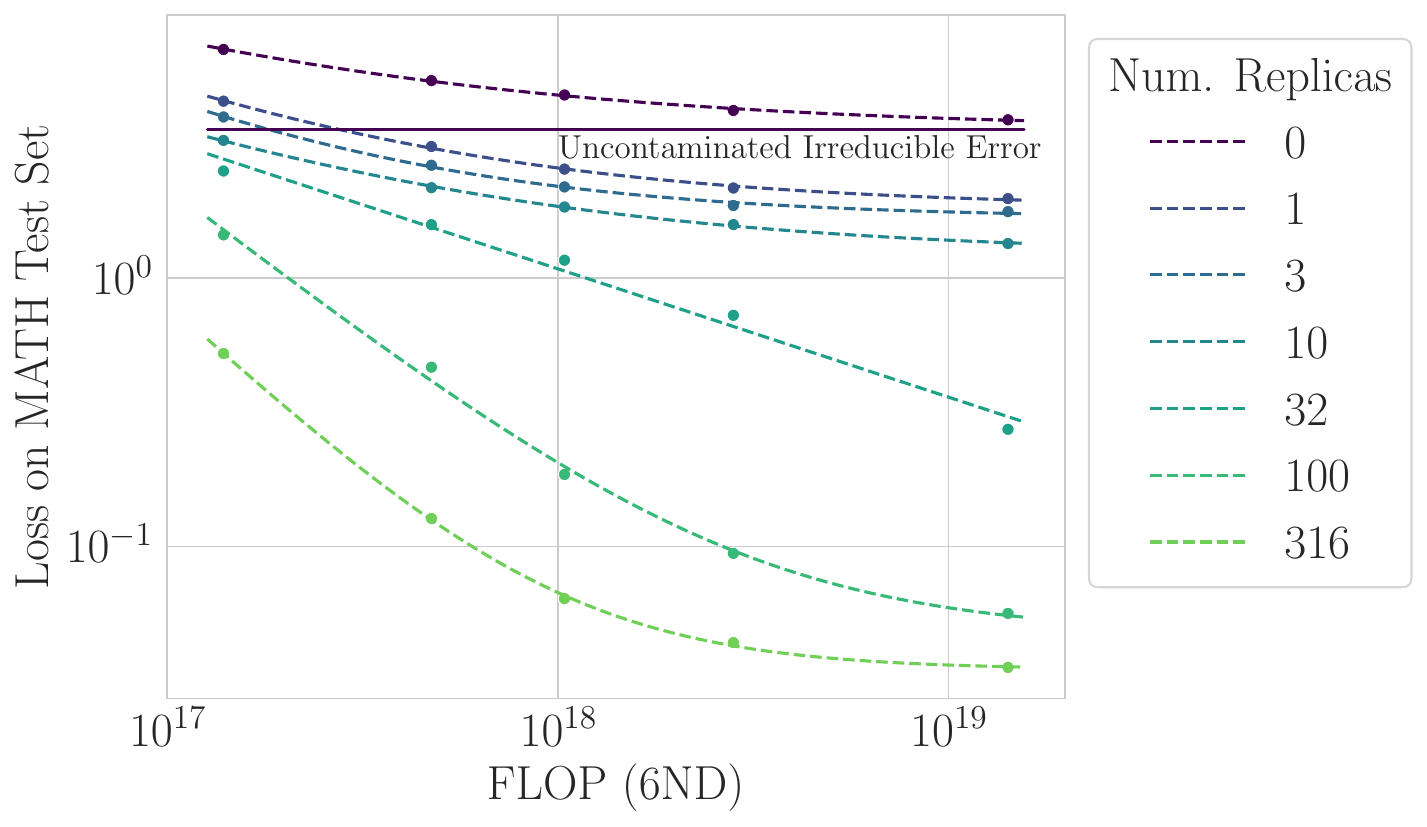}
    \includegraphics[width=\linewidth]{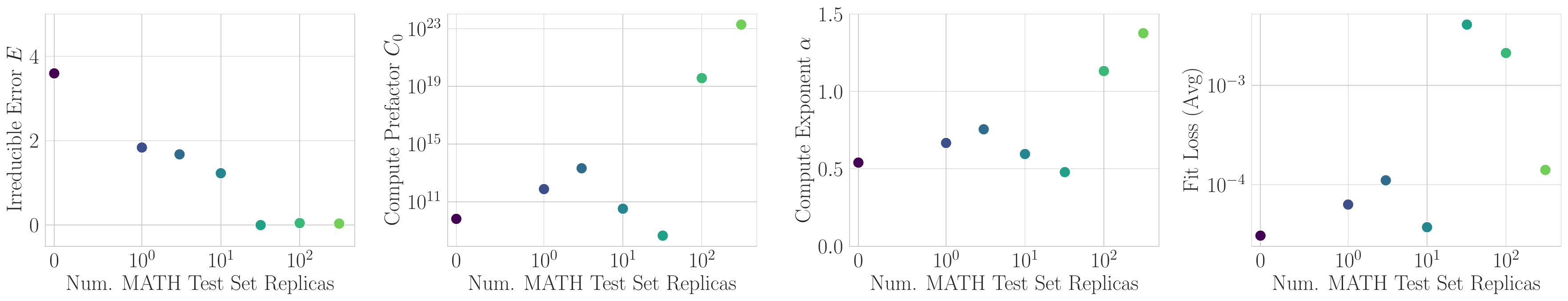}
    \caption{\textbf{Scaling Laws Suggest Including A Single Test Set Replica Achieves Lower Loss Than the Irreducible Error of the Uncontaminated Corpus.} \textbf{Top:} For each scaling series contaminated with $R$ replicas of the MATH test set, we fit scaling laws $\mathcal{L}(C, R) = E(R) + C_0(R) \cdot C^{-\alpha(R)}$, where $C \approx 6 N D$ is pretraining compute.
    Almost all contaminated models achieve lower cross entropy on the test set than the irreducible error of training on uncontaminated data (horizontal purple line).
    \textbf{Bottom:} Increasing test set contamination reduces the irreducible error $E(R)$ from $3.594$ at $R=0$ to $0.0347$ at $R=316$. Larger values of $R$ also increase the compute prefactor and compute exponent.
    The functional form achieves average fitting error $< 10^{-2}$ for all $R$.
    }
    \label{fig:contamination_scaling_laws}
\end{figure*}

In this work, we quantitatively study the effects of test set contamination on generative evaluations, focusing on the widely used MATH benchmark \citep{hendrycks2021measuringmathematicalproblemsolving}.
We pretrain dozens of language models on corpora contaminated with varying numbers of test set replicas, then study the effects of contamination through the language model lifecycle:

\textbf{Pretraining: Scaling \& Irreducible Error} (Sec.~\ref{sec:pretraining}). We find that performance increases with contamination and model size, similar to discriminative settings. Under standard scaling law assumptions, we discover a fundamental breach: including even a single replica of the test set enables achieving a lower loss than the irreducible error of the uncontaminated corpus.

\textbf{Subsequent Training: Overtraining \& Supervised Finetuning} (Sec.~\ref{sec:further_training}). We find that training beyond compute-optimal with fresh data dilutes the performance gains from contamination. We further discover that Supervised Finetuning (SFT) has opposing effects depending on pretraining contamination: it improves (worsens) performance for low (high) contamination.

\textbf{Deployment: Temperature \& Solution Length} (Sec.~\ref{sec:inference}) Sampling temperature and solution length modulate memorization, leading us to mathematically characterize three regimes: Exponentially Fast Decoherence, Brittle Memorization, and Deterministic Lock-In. We show that high-temperature sampling acts as a "truth serum," decoupling contamination gains from generalization.

As an final and general contribution, we discover and fix a critical implementation error in the EleutherAI LM Evaluation Harness \citep{gao2024evalharness} for Math Verify Scores (Appendix~\ref{app:sec:math_verify_bug}), ensuring that valid reference solutions are accurately scored as correct (raising scores of gold references from $\mathord{\sim}$70\% to 100\%), a necessary step for trustworthy reporting on mathematical benchmarks.

\section{Methods}
\label{sec:methods}

\textbf{Pretraining} We pretrained causal language models \citep{vaswani2017attention, brown2020language} using the Qwen 3 architecture \citep{yang2025qwen3technicalreport}, sweeping model sizes: $34$M, $62$M, $93$M, $153$M, $344$M.
For each model size, we created multiple pretraining corpora by contaminating a high quality corpus \citep{penedo2024finewebdatasetsdecantingweb} with a different number of replicas of the benchmark test set: from $0$ (uncontaminated) through $1, 3, 10, 32, 100, 316, 1000, 3162$ (logarithmically uniform).
Pretraining floating point operations (FLOP) $C$ was approximated $ 6  \, N \, D$ \citep{kaplan2020scaling,sardana2024beyond,porian2024resolving,gadre2024languagemodelsscalereliably}, where $N$ is model parameters and $D$ is tokens.
Each model was pretrained on 20 tokens-per-parameter \citep{hoffman2022chinchilla}.
For each tuple of parameters and test set replicas, we trained one model.
For details, see App.~\ref{app:sec:pretraining_implementation_details}.

As a key distinction, \citet{bordt2025howmuch} pretrains models up to ~1.6B, but achieves this by contaminating disjoint subsets of the test set at different replica counts within a single training run --- a cheaper but messier design. We pretrain separate models per contamination amount for tighter experimental control, which is significantly more expensive and thus limits our maximum scale.

\textbf{Benchmark} We chose the MATH \citep{hendrycks2021measuringmathematicalproblemsolving} benchmark of competition math problems because it is comparatively large (5,000 test problems), and includes solutions as well as verifiable answers.
The MATH test set contains $\mathord{\sim}1.4$M tokens under the Qwen 3 tokenizer.

\textbf{Evaluation}
The first metric we report is \emph{Math Verify} \citep{kydlicek2025fixing}, defined as the fraction of problems for which the model generates solutions verified to be mathematically equivalent to the boxed answers.
We initially used EleutherAI's LM Evaluation Harness \citep{gao2024evalharness}, but discovered a bug in how Math Verify is computed; for example, the benchmark's gold reference solutions obtained a score of $\mathord{\sim}70\%$.
We worked with the developers to correct the implementation (Appendix~\ref{app:sec:math_verify_bug}).
This suggests that \texttt{math\_minerva} scores using the Eval Harness prior to Aug 23, 2025 may be incorrect.
The second metric we report is the \emph{Cross Entropy} of the gold references, which was previously demonstrated to be useful for studying scaling properties of generative evaluations during pretraining \citep{schaeffer2025pretrainingscalinglawsgenerative}.
We used temperature-only sampling \citep{schaeffer2025minpmaxexaggerationcritical}, beginning with temperature $0$, and expanding to more temperatures in Sec.~\ref{sec:inference}.

\section{Pretraining: Scaling \& Irreducible Error}
\label{sec:pretraining}

We begin at the first stage of the model lifecycle: pretraining.

\paragraph{Finding \#1: Performance Increases with Contamination and Model Size}
Consistent with discriminative evaluations, adding benchmark replicas to the pretraining corpus increases Math Verify scores and decreases cross entropies (Fig.~\ref{fig:math_verify_ce_temp_0} Left), as does increasing model size (Fig.~\ref{fig:math_verify_ce_temp_0} Right). The relationship between replicas and performance is non-linear: at low contamination ($\leq 10$ replicas), models perform near the baseline, but at $\mathord{\sim}100$ replicas, performance jumps to ceiling.

\begin{table}[t!]
    \centering
    \begin{minipage}{0.32\textwidth}
        \centering
        \begin{tabular}{crr}
            \toprule
            \textbf{Replicas} & \textbf{Reph.} & \textbf{Pert.} \\
            \midrule
            0    & 0.04\% & 0.00\% \\
            1    & 0.00\% & 0.00\% \\
            3    & 0.00\% & 0.00\% \\
            10   & 0.00\% & 0.00\% \\
            32   & 0.00\% & 0.00\% \\
            100  & 0.00\% & 0.00\% \\
            316  & 0.02\% & 0.00\% \\
            \phantom{0}\\
            \phantom{0}\\
            \bottomrule
        \end{tabular}
        \subcaption*{34M}
    \end{minipage}\hfill
    \begin{minipage}{0.32\textwidth}
        \centering
        \begin{tabular}{crr}
            \toprule
            \textbf{Replicas} & \textbf{Reph.} & \textbf{Pert.} \\
            \midrule
            0    & 0.04\% & 0.00\% \\
            1    & 0.00\% & 0.00\% \\
            3    & 0.00\% & 0.00\% \\
            10   & 0.00\% & 0.00\% \\
            32   & 0.00\% & 0.00\% \\
            100  & 0.00\% & 0.00\% \\
            316  & 0.00\% & 0.00\% \\
            \phantom{0}\\
            \bottomrule
        \end{tabular}
        \subcaption*{93M}
    \end{minipage}\hfill
    \begin{minipage}{0.32\textwidth}
        \centering
        \begin{tabular}{crr}
            \toprule
            \textbf{Replicas} & \textbf{Reph.} & \textbf{Pert.} \\
            \midrule
            0    & 0.04\% & 0.00\% \\
            1    & 0.00\% & 0.00\% \\
            3    & 0.00\% & 0.00\% \\
            10   & 0.00\% & 0.00\% \\
            32   & 0.00\% & 0.00\% \\
            100  & 0.02\% & 0.00\% \\
            316  & 0.00\% & 0.00\% \\
            \bottomrule
        \end{tabular}
        \subcaption*{344M}
    \end{minipage}
    \caption{\textbf{Contamination-Driven Performance Is Memorization, Not Generalization.} When MATH test problems are rephrased (same numbers, different wording) or perturbed (same wording, different numbers), model performance collapses to baseline regardless of contamination level or model size, confirming that gains from contamination are not attributable to general mathematical reasoning. Results were consistent across all model sizes.}
    \label{tab:rephrase_perturb}
\end{table}

\paragraph{Finding \#2: Contamination-Driven Performance Is Memorization, Not Generalization}
To determine whether performance gains stemmed from generalization or memorization, we evaluated contaminated models on two modified versions of the MATH test set. First, we \emph{rephrased} the problems, keeping the original numerical values and logic but altering the linguistic surface form. Second, we \emph{perturbed} them, modifying numerical values and correct answers while preserving the problem structure. In both conditions, performance regresses to match uncontaminated models across all model sizes and contamination levels (Table~\ref{tab:rephrase_perturb}). This strongly suggests that gains from contamination rely on verbatim memorization rather than acquired mathematical reasoning.

\paragraph{Finding \#3: Scaling Laws Suggest Including A Single Test Set Replica Achieves Lower Loss Than the Irreducible Error of the Uncontaminated Corpus}

How much does test set contamination ``buy" a model?
More specifically, how much pretraining compute must be spent on an uncontaminated pretraining corpus to match the performance of a model trained on a corpus containing $R$ replicas of the benchmark test set?
We answered this via scaling laws \citep{kaplan2020scaling, hoffman2022chinchilla, openai2024gpt4technicalreport, hu2024predictingemergentabilitiesinfinite, schaeffer2025pretrainingscalinglawsgenerative}. For each scaling series pretrained on corpora contaminated with $R$ replicas of the MATH test set, we fit neural scaling laws for the cross entropy loss $\mathcal{L}$ on the benchmark test set as a function of pretraining compute $C$:
\begin{equation}
    \label{eqn:neural_scaling_law}
    \mathcal{L}(C, R) = E_0(R) + C_0(R) \cdot C^{-\alpha(R)}.
\end{equation}
Fig.~\ref{fig:contamination_scaling_laws} shows each scaling law, as well as each scaling law's estimated parameters.
Our models' pretraining compute budgets and losses on the test set are reasonably well fit.
The biggest effect of increasing contamination is that the irreducible error shrinks from $E = 3.594 \rightarrow 0.0347$ as $R=0 \rightarrow 316$.
On the models we trained, \emph{including even a single replica enables almost all models to achieve lower cross entropy losses than the estimated irreducible error of the uncontaminated pretraining corpus}.
Under standard scaling law assumptions -- specifically, that the scaling law extrapolates infinitely -- a contaminated pretraining corpus can buy more than an ``infinite'' amount of compute relative to pretraining on an uncontaminated pretraining corpus.

This conclusion potentially contradicts \citet{huang2024demystifying}'s claim that single-shot verbatim memorization is an ``illusion''  and \citet{hayes2025exploring}'s claim that membership inference attacks are limited on pre-trained LLMs.
Future work should investigate this difference.

\begin{figure*}[t!]
    \centering
    \includegraphics[width=\linewidth]{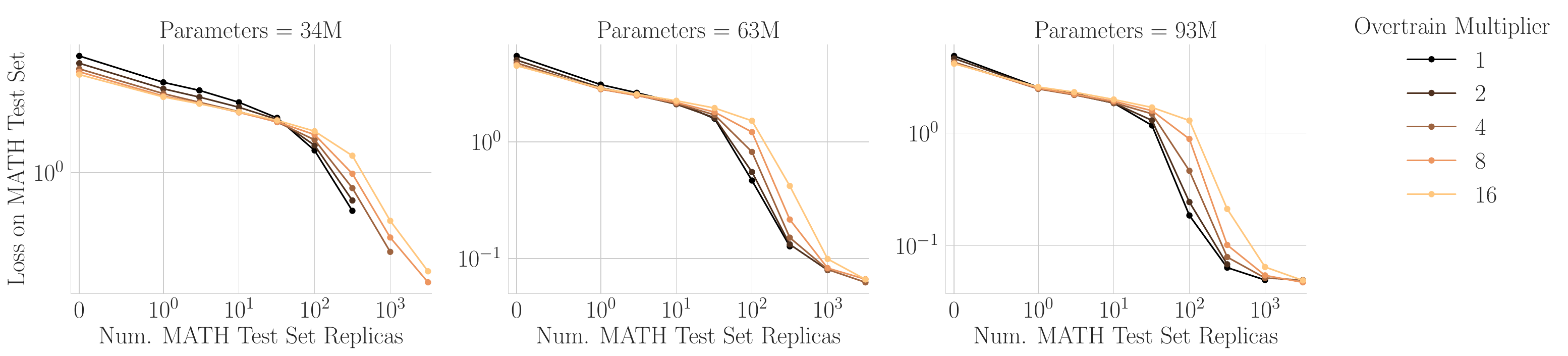}
    \includegraphics[width=\linewidth]{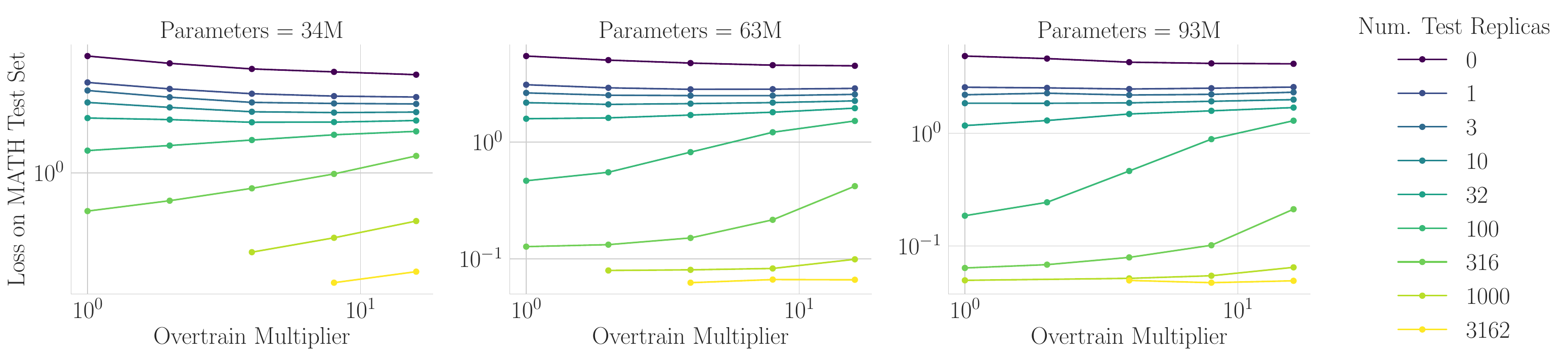}
    \caption{\textbf{Overtraining with Fresh Data Mitigates Contamination.} Consistent with discriminative evaluations \citep{bordt2025howmuch}, overtraining (training longer than Chinchilla compute-optimal) on \emph{new fresh data} interacts with contamination: increasing overtraining decreases cross entropy on the MATH test set for uncontaminated models, but increases cross entropy for contaminated models. This suggests that while fresh data generally improves performance, it dilutes the ``dose,'' or proportion of contaminated pretraining tokens, thereby lessening the effect of the contamination. The crossover point shifts with model size, falling from 32 test set replicas for 34M to 1 replica for 93M models, indicating larger models lose their contamination advantage more readily when overtrained.}
    \label{fig:overtraining}
\end{figure*}

\section{Further Training: Overtraining \& Supervised Finetuning}
\label{sec:further_training}

We next examine how two common subsequent training stages---overtraining on fresh data and supervised finetuning---interact with pretraining contamination.

\paragraph{Finding \#4: Overtraining with Fresh Data Mitigates Contamination}
\citet{bordt2025howmuch} reported that for discriminative evaluations, the effect of contamination diminishes when models are trained beyond ``compute optimal'' on fresh data. We tested whether this so-called \emph{overtraining} \citep{touvron2023llama2openfoundation,sardana2024beyond,gadre2024languagemodelsscalereliably,schaeffer2025pretrainingscalinglawsgenerative} has similar effects for generative evaluations.
We extended our Sec.~\ref{sec:methods} sweep into the overtrained regime, pretraining on
$D(m, N) \defeq m \times 20 \times N$
tokens per model, where $m$ is the \emph{overtraining multiplier} and $N$ is the number of model parameters.
We swept $m \in \{1, 2, 4, 8, 16\}$.
We term $m=1$ ``compute-optimal training'' and $m > 1$ ``overtraining'' \citep{sardana2024beyond,gadre2024languagemodelsscalereliably}.
Crucially, as the overtraining multiplier increases, the additional tokens are \emph{new, fresh tokens}; this differs from settings where models might see select documents repeated tens-to-hundreds of times \citep{hernandez2022scalinglawsinterpretabilitylearning} or the entire corpus repeated for 4+ epochs \citep{muennighoff2023scaling,fang2025datasets}.

\begin{figure*}[t!]
    \centering
    \includegraphics[width=1.0\linewidth]{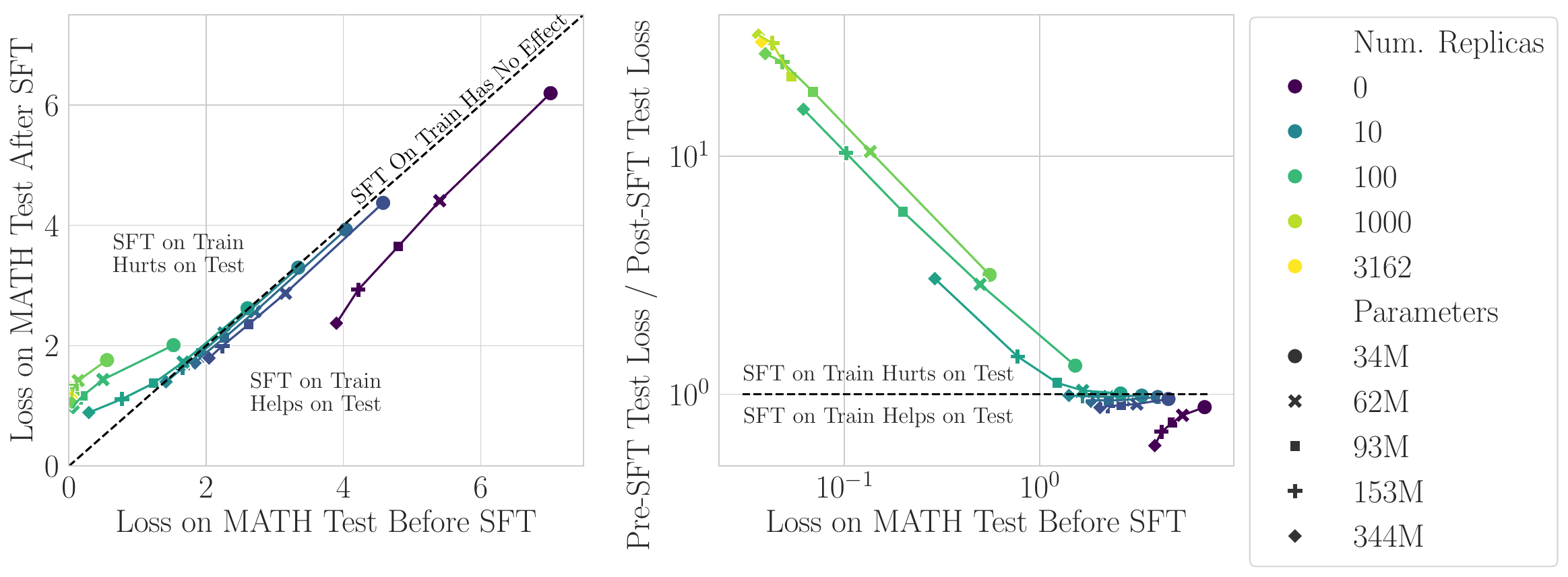}
    \caption{\textbf{Supervised Finetuning on the Train Set Has Opposing Effects, Depending on Pretraining Contamination.} For little-to-no contamination ($<10$ test set replicas), supervised finetuning (SFT) on the MATH \emph{train} set decreases loss on the \emph{test} set. For models pretrained with more contamination ($>10$ replicas), SFT on the train set increases loss on the test set.
    }
    \label{fig:sft}
\end{figure*}

For models with low contamination, increasing overtraining decreases cross entropy on the MATH test set, but for models with high contamination, overtraining increases the cross entropy (Fig.~\ref{fig:overtraining}).
The cross-over point between test set replicas and overtraining multiplier shifts with model size: it falls from 32 test set replicas for $34$M models to 1 replica for $93$M models.
Thus, as models become larger, the performance boost from contamination diminishes when overtraining with fresh data.
While more fresh data is generally useful for improving model performance, it dilutes the ``dose'' of the contaminated data, weakening benchmark scores \citep{schaeffer2025doseresponse, wei2025hubblemodelsuiteadvance}.

\paragraph{Finding \#5: Supervised Fine-Tuning on Training Set Has Opposing Effects, Depending on Pretraining Contamination}

After pretraining, the first post-training step is oftentimes supervised finetuning (SFT) \citep{wei2022finetunedlanguagemodelszeroshot, ouyang2022training}. We turned to assessing what effect SFT has on contaminated pretrained models. \citet{kocyigit2025the} recently found that SFT on the \emph{train} set improves performance on the \emph{test} set. However, as a key methodological difference, \citet{kocyigit2025the} induced test set contamination via continued pretraining \citep{jin2022lifelong,jang2022continualknowledgelearninglanguage,ibrahim2024simple, parmar2024reusedontretrainrecipe,yildiz2025investigatingcontinualpretraininglarge}, whereas we introduced contamination via pretraining.
The MATH format was preserved between pretraining and SFT.

\begin{figure*}[t!]
    \centering
    \includegraphics[width=\linewidth]{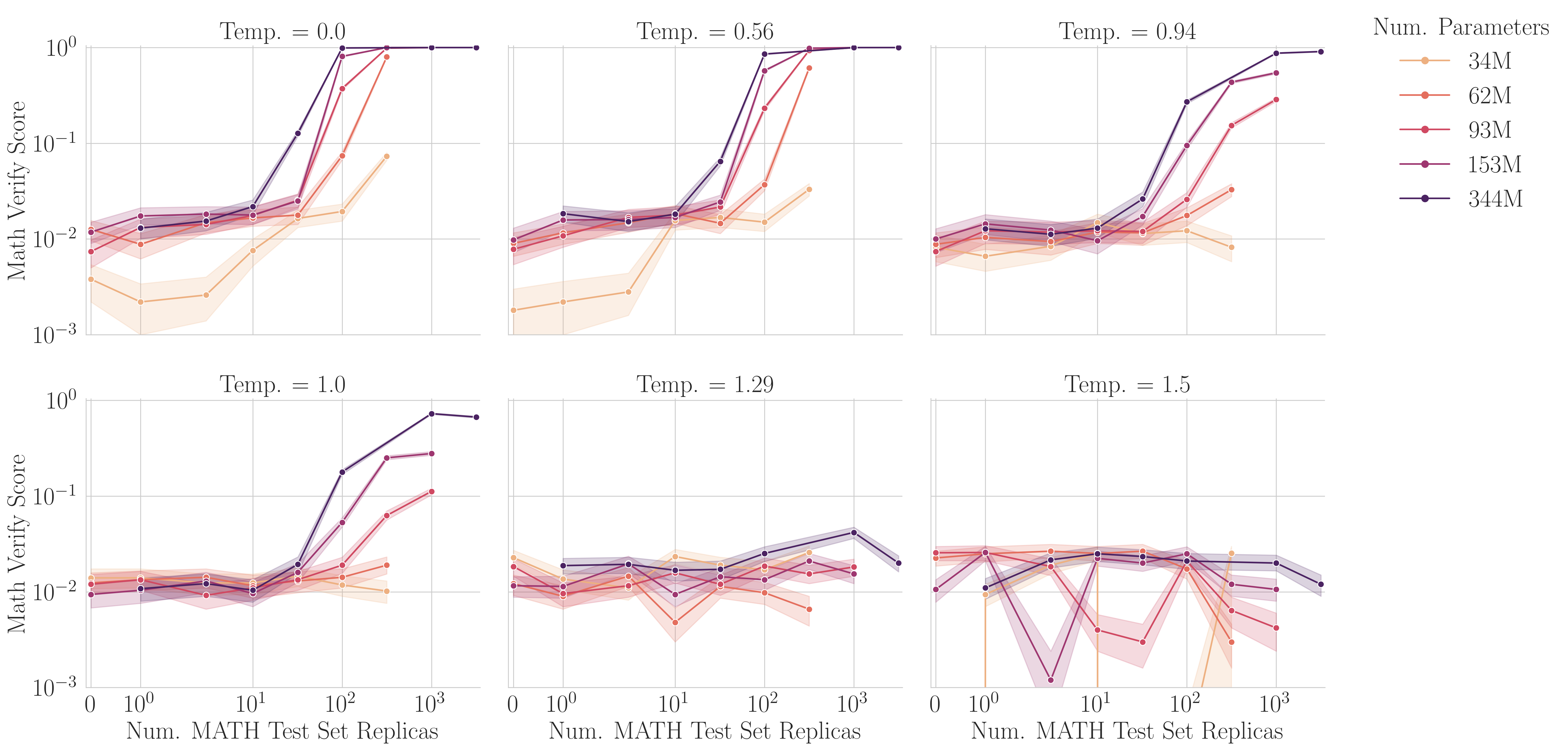}
    \caption{\textbf{Sampling Temperature Degrades Performance, Particularly for Contaminated Models.}
    As sampling temperature increases, Math Verify scores drop, falling from near 100\% to under 1\% in many configurations.
    The drop is disproportionately larger for highly contaminated models: increasing temperature from 0 to 1 reduces performance by a factor of $\sim2$ at low contamination ($\leq 10$ replicas), whereas it reduces performance by a factor of up to 40 at high contamination.
    }
    \label{fig:math_verify_replicas_temp}
\end{figure*}

\begin{figure*}[t!]
    \centering
    \includegraphics[width=\linewidth]{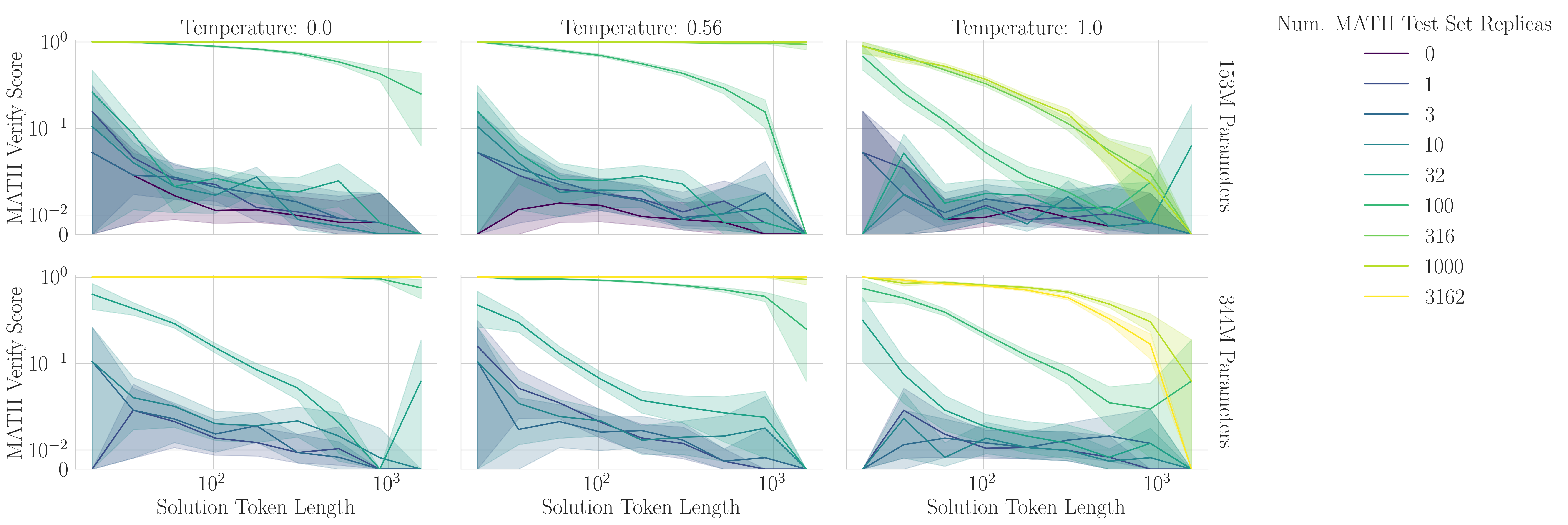}
    \caption{\textbf{Performance Declines with Increasing Solution Token Length.} Math Verify Scores decrease with increasing solution length, modulated by temperature.  At high rates of contamination and higher temperatures, Math Verify scores fall exponentially with the solution length.  At lower temperatures, and high rates of contamination, solution length has almost no effect.  Models exposed to less contamination also exhibit declining scores with longer solution lengths, but the dropoff is concave up and occurs much more slowly.
    }
    \label{fig:temp_and_solution_length}
\end{figure*}

We discover that, surprisingly, SFT on the train set can help or hurt performance on the test set, depending on the amount of pretraining contamination (Fig.~\ref{fig:sft}).
For models with no or low $(< 10)$ test set replicas, SFT on the train set significantly reduces loss on the test set.
At $10$ replicas, SFT has no effect, but as contamination increases, SFT on the train set significantly increases loss on the test set.
Teacher-forced evaluation of SFT checkpoints on perturbed MATH (novel problems never seen during pretraining or SFT) shows SFT lowers NLL at 14/17 conditions, with improvements up to -2.8 nats; this reveals that during SFT, contaminated models learn to generalize, but also forget their contaminated pretraining data, and the effects of contamination are more impactful than generalization for small models, leading to a counter-intuitive net increase in test loss.

\begin{figure*}[t!]
    \centering
    \includegraphics[width=1.0\linewidth]{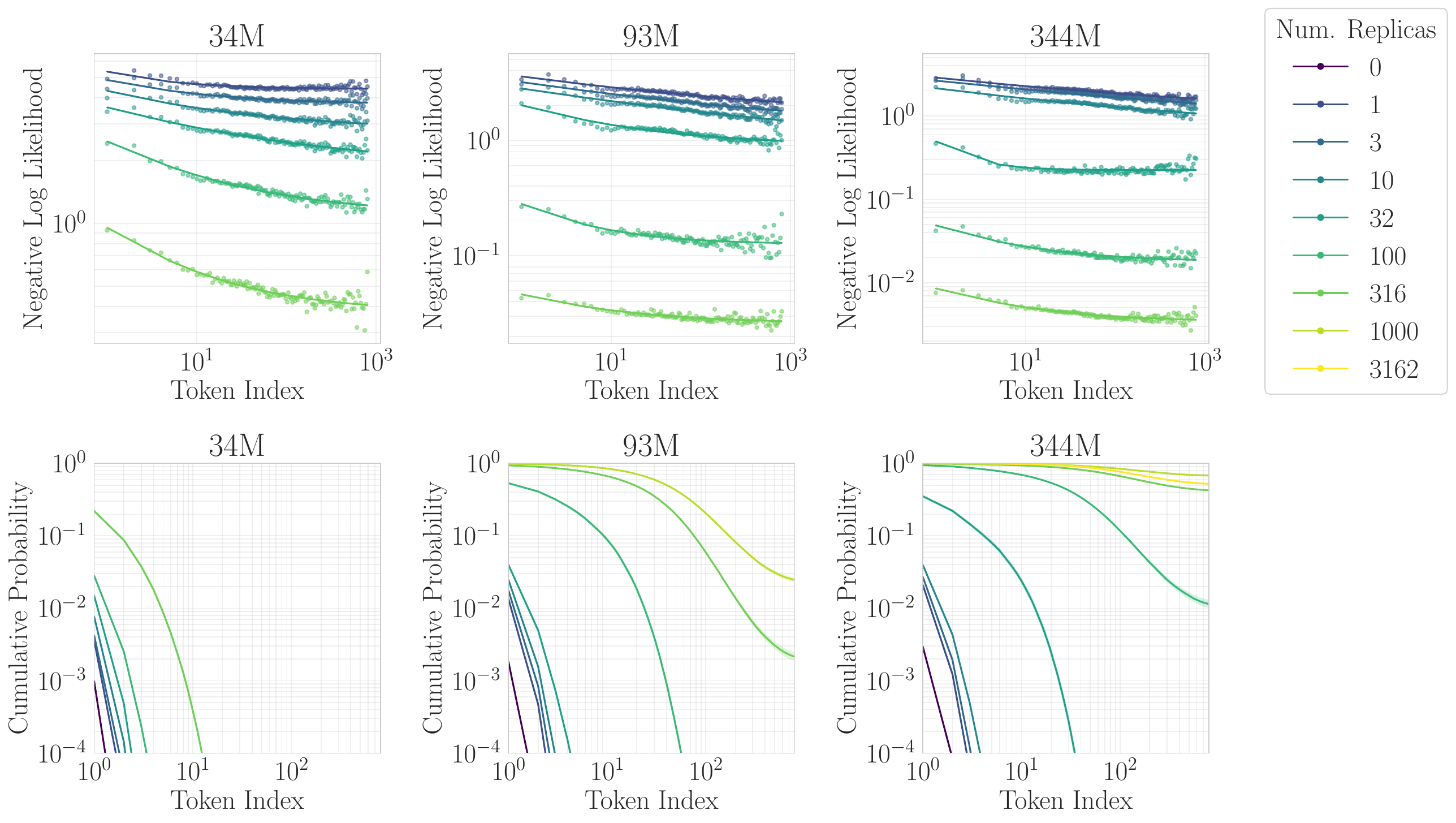}
    \caption{\textbf{Solution Length and Temperature Together Govern How Quickly Memorization Decoheres.}
    \textbf{Top:} Most models and contamination pairs exhibit scaling laws as a function of sequence length (Eqn.~\ref{eqn:in_context_scaling_law}). \textbf{Bottom:} The cumulative probability of generating a memorized solution lives in one of three regimes: (1) Exponentially Fast Decoherence, (2) Brittle Memorization, and (3) Deterministic Lock-In. Sampling temperature shifts the scaling exponent, and can shift a model between regimes.
    }
    \label{fig:survival_process}
\end{figure*}

\section{Inference: Temperature \& Solution Length}
\label{sec:inference}

Finally, we arrive at the last stage of the model lifecycle, deployment, where sampling and the sequence length of generations are new considerations absent from discriminative evaluations.

\paragraph{Finding \#6: Increasing Temperature Mitigates Gains from Contamination}

We evaluated the pretrained models using temperature-only sampling from $\tau = 0$ (``greedy'') to $1.5$ (Fig.~\ref{fig:math_verify_replicas_temp}).
Math Verify scores remain stable for low-temperature sampling ($\tau \leq 0.56$).
Beyond this point, performance degrades quickly: increasing temperature to $1$ causes performance at high contamination levels ($1000$ replicas) to collapse by $40\times$, down to performance of uncontaminated models.
Thus, small adjustments in inference can eliminate the benefits of contamination entirely.

\paragraph{Finding \#7: Longer Solution Length Mitigates Gains from Contamination}

To understand how solution length constrains performance, we binned problems into 10 log-spaced intervals ranging from the shortest (15 tokens) to the longest (1949 tokens).
Math Verify scores decrease significantly as solution length increases (Fig.~\ref{fig:temp_and_solution_length}).
We identified a striking shift in the functional form of this decay based on contamination level:
At higher contamination levels, the decay follows an approximate power law, but at lower levels of contamination, performance decays exponentially quickly to the noise floor.
This suggests that maintaining a coherent memorized chain becomes increasingly difficult as the sequence grows, which aligns with findings that longer sequences require more repetitions to be memorized \citep{jiang2025a,lu2024scaling}.

Furthermore, solution length interacts with temperature:
For short solutions ($\leq 100$ tokens) at high contamination (316 replicas), raising the temperature from $0$ to $1.0$ drops accuracy by $\mathord{\sim}45\%$ on the largest model; for $400$ tokens, the same temperature increase causes accuracy to drop by $\mathord{\sim}100\%$.

\paragraph{Finding \#8: Temperature and Solution Length Together Govern the Survival Process}

Generative evaluations differ fundamentally from discriminative tasks because they require the model to maintain a coherent trajectory over a sequence of length $T$.
For models that solve the task via memorizing solutions, we can characterize how the model must ``survive'' the risk of decoherence at every token step $t$ via studying the probabilities under teacher forcing.

\begin{figure}[t!]
    \centering
    \includegraphics[width=0.9\columnwidth]{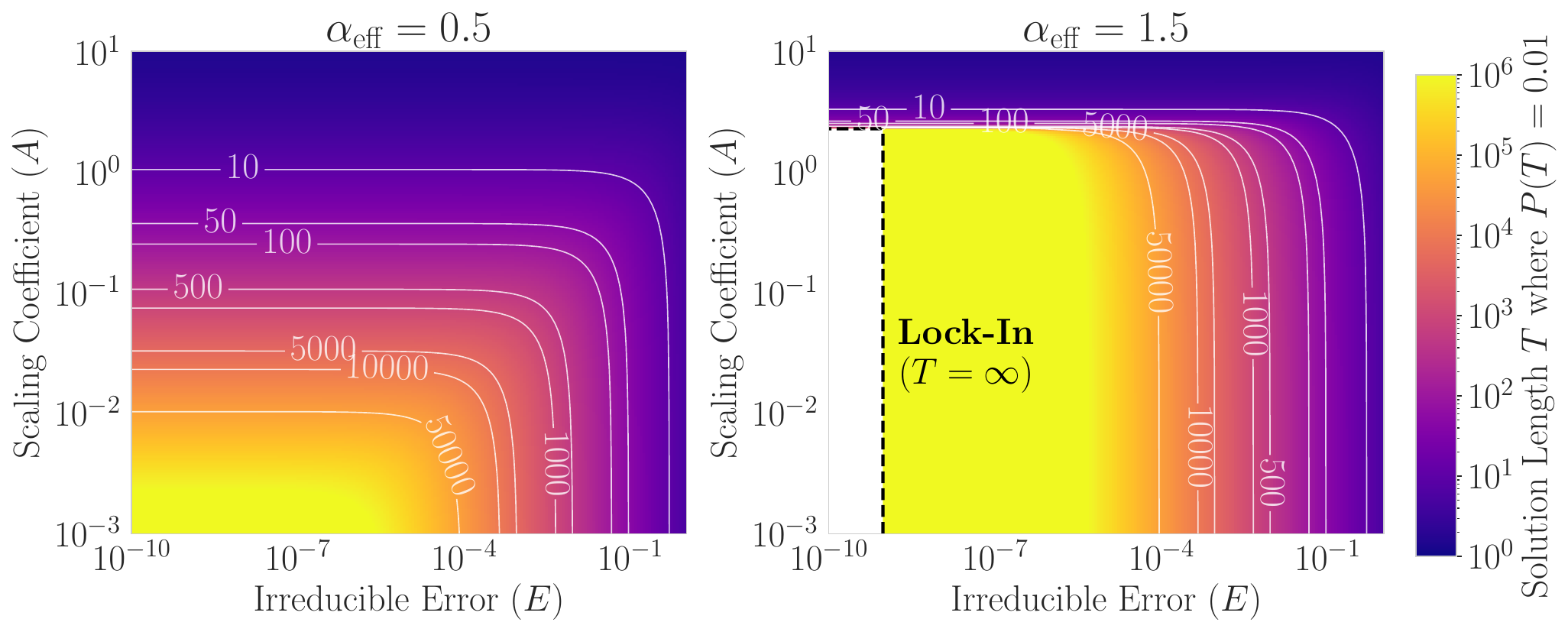}
    \caption{\textbf{Regimes of Memorization.}
    Phase diagrams illustrating the maximum sustainable solution length $T$ for a fixed survival probability $P(T) = 0.01$. The three regimes are (1) Exponentially Fast Decoherence, (2) Brittle Memorization, and (3) Deterministic Lock-In.}
    \label{fig:phase_diagram}
\end{figure}

Motivated by prior work on in-context scaling laws~\citep{anthropic2023claude2,geminiteam2024gemini15unlockingmultimodal,xiong2024effective,anil2024many}, we find that for most model sizes and number of test set replicas, the negative log likelihood scales with the token index $t$ as a power law plus an irreducible error (Fig.~\ref{fig:survival_process}):
\begin{equation}\label{eqn:in_context_scaling_law}
    \mathcal{L}_t(N, R) \approx E(N, R) + A(N, R) \cdot t^{-\alpha(N, R)}.
\end{equation}
However, we do find a small number of notable exceptions (Appendix.~\ref{app:sec:exceptions_to_sequence_scaling_laws}).
The probability of successfully generating a solution of length $T$ can then be approximated as:
\begin{equation}
    \label{eq:survival_prob}
    P(T) \approx \exp\left( -E(T-1) - \tfrac{A}{1-\alpha}(T^{1-\alpha} - 1) \right).
\end{equation}
Intuitively, the probability of generating the solution sequence decays exponentially with sequence length, but the probability of predicting the correct next token increases exponentially with sequence length.
These two opposing forces yield three distinct memorization regimes (Fig.~\ref{fig:phase_diagram}):

\paragraph{Regime I: Exponentially Fast Decoherence ($E > 0$).}
When the irreducible error term $E$ dominates, the survival probability decays exponentially ($P(T) \sim e^{-E \cdot T}$), rendering the generation of long solutions statistically improbable.

\paragraph{Regime II: Brittle Memorization ($E \approx 0, \alpha \le 1$).}
At intermediate contamination, the probability follows a Weibull-like stretched exponential ($P(T) \sim e^{-k T^{1-\alpha}}$). While the model can accurately generate sequences, the cumulative probability of success eventually vanishes.

\paragraph{Regime III: Deterministic Lock-In ($E \approx 0, \alpha > 1$).}
At high contamination, $\alpha > 1$ flips the sign of the exponent in Eq.~\ref{eq:survival_prob}, causing the penalty term $T^{1-\alpha}$ to decay to zero and $P(T)$ converges to a non-zero constant even as $T \to \infty$. The model's certainty increases faster than error accumulates, enabling \textit{deterministic lock-in} of long solutions.

\paragraph{The Effect of Sampling Temperature.}
While the native scaling $\alpha$ is a property of the trained model, inference behavior is modulated by the sampling temperature $\tau$.
Since temperature acts as a scalar divisor on logits ($z \to z/\tau$), it modifies the rate at which probability mass concentrates on the memorized path, yielding an \textit{effective} scaling exponent:
$\alpha_{\mathrm{eff}}(\tau) \approx \alpha / \tau$.
This reveals that temperature can artificially shift a model between regimes, exposing the fragility of memorization: a model operating in Deterministic Lock-In ($\alpha_{\mathrm{eff}} > 1$) at greedy settings ($\tau \to 0$) can be reverted to Brittle Memorization ($\alpha_{\mathrm{eff}} < 1$) by increasing $\tau$.
However, sampling at $\tau=1.0$ exposes the true exponent $\alpha < 1$, causing failure for long sequences.
This explains why high-temperature sampling acts as a ``truth serum,'' decoupling the gains of contamination from robust generalization.

\section{Discussion}

Benchmarks serve as our proxies for AI capabilities in the wild; contamination breaks this proxy, creating an illusion of competence.
This work provides the first targeted examination of contamination in generative tasks.
Our scaling laws reveals the magnitude of this threat. Under standard scaling law assumptions, a single replica of the test set allows a model to bypass the estimated ``irreducible error'' barrier, achieving performance that would otherwise require more than an infinite amount of compute on uncontaminated data.
However, this performance is brittle; critical ``levers'' -- sampling temperature and solution length -- differentiate memorization from generalization.

Perhaps most critically for practitioners, training pipelines can mask or expose this issue in counter-intuitive ways. The interaction between contamination and further training is governed by a ``dose-response'' effect \citep{schaeffer2025doseresponse, wei2025hubblemodelsuiteadvance}: overtraining on fresh data dilutes the contamination, and SFT on training data can degrade test performance by overwriting memorized test problems.
This implies that a drop in test accuracy after SFT, usually a sign of alignment tax or forgetting, may actually be a positive signal of decontamination.

\paragraph{Limitations.}
Our study isolates contamination on MATH, a single automatically verifiable generative benchmark, using decoder-only dense Qwen 3 models up to 344M parameters and one web-crawl pretraining mixture.
These choices make the causal study controlled, but they limit direct extrapolation to larger frontier models, other architectures, and other generative domains.
In particular, 344M parameters is small relative to frontier scales, where contamination and genuine reasoning capability would interact in ways our experiments cannot directly observe; our scaling-law analysis (Finding~\#3) provides a formal bridge to larger compute budgets, though specific quantitative thresholds may shift at scale.
Similarly, restricting evaluation to MATH leaves cross-benchmark generalization untested; however, the dynamics we characterize derive from properties of sequential generation rather than MATH specifically -- temperature sensitivity, solution-length decay, and the survival-process framework apply to any generative task whose per-token NLL admits an in-context scaling law -- and extension to code and reasoning benchmarks is valuable future work.
Appendix~\ref{app:sec:discussion} discusses additional limitations and future directions.

\begin{ack}
RS acknowledges support from Stanford Data Science and from the OpenAI Superalignment Fast Grant.
JK acknowledges support from NSF grant number DGE1656518.
SK acknowledges support by NSF 2046795 and 2205329, the MacArthur Foundation, Stanford HAI, OpenAI and Google Inc.
KZL is generously supported by the Amazon AI PhD Fellowship.
AA is generously supported by a Knight-Hennessy Fellowship, an NSF Graduate Research Fellowship, and a Georgetown Foundation Research Grant.
Table~\ref{tab:rephrase_perturb} experiments were run using a computing grant provided by TensorPool.
\end{ack}

\clearpage

\bibliographystyle{preprintnat}
\bibliography{references_rylan}

\appendix

\section{Related Work}
\label{app:sec:related_work}

\paragraph{Data Contamination and its Consequences}
Test set contamination, where benchmark data is included in pretraining corpora, is widely recognized as a threat to valid model evaluation, as it can lead to inflated performance metrics. Numerous survey and position papers have documented the various ways contamination can occur and have called for routine audits and transparent reporting for all benchmarks \citep{sainz2023nlp, sainz2024data, deng2024unveiling, xu2024benchmarkdatacontaminationlarge, reuel2025open}. Empirical studies of large web-scale datasets have confirmed significant overlap and duplication between training and test sets \citep{dodge2021documenting}. Research focused on ensuring benchmark integrity has identified multiple ways that language models might "cheat" on evaluations if contamination is not properly managed \citep{zhou2023dontmakellmevaluation, dong2024generalizationormemorization}. For instance, analyses of popular mathematics benchmarks have revealed signals of data leakage and potential overfitting \citep{zhang2024carefulexaminationlargelanguage}. Ongoing community efforts and open-source audits continue to measure the extent of contamination across different models and datasets \citep{li2024opensource}. The risks extend beyond evaluation integrity; scaling studies indicate that poisoning risks increase with model size, as larger models learn harmful behaviors from minuscule amounts of poisoned data far more rapidly than smaller models, underscoring the necessity of robust data curation \citep{bowen2025scalingtrendsdatapoisoning}. As a cautionary illustration, \citet{schaeffer2023pretrainingtestsetneed} demonstrated that pretraining on the test set is a trivial path to strong benchmark performance, reinforcing the importance of rigorous decontamination and auditing.

\paragraph{Controlled Contamination During Pretraining}

A line of research directly investigates the causal effects of contamination by intentionally adding benchmark data to pretraining corpora and observing the results. \citet{magar2022data} interleaved task-specific datasets into a general text corpus during pretraining, varying the duplication rate of the leaked examples. They differentiated between "memorization" (storing examples) and "exploitation" (using stored examples to boost test scores), finding that both model size and the number of repetitions increased exploitation. \citet{jiang2024investigatingdatacontaminationpretraining} pretrained models from scratch on corpora containing either only the inputs ("text-only") or the full input-output pairs ("ground-truth") of benchmark examples, sweeping the contamination frequency. They observed significant performance gains when ground-truth pairs were used and showed that simple n-gram-based detection methods could be bypassed by paraphrasing or partial data leaks. The problem also transcends language barriers; \citet{yao2024data} demonstrated a cross-lingual contamination channel where continuing to pretrain a model on non-English translations of English benchmarks led to material improvements on the original English tests, a form of contamination that string-matching would not detect. At a larger scale, \citet{bordt2025howmuch} varied the repetition count of leaked examples, model size (up to ~1.6B parameters), and the total training token budget, finding that performance scales predictably with size and repetition. They also showed that sufficiently long training on abundant unique data could mitigate or even reverse the effects of earlier contamination. In the context of machine translation, \citet{kocyigit2025overestimation} injected source-target pairs into the pretraining data of 1B and 8B parameter models, quantifying significant overestimation in BLEU scores, with larger models and low-resource languages showing more pronounced effects. Together, these causal intervention studies provide clear evidence that language models memorize and leverage benchmark data when it is present during pretraining.

\paragraph{Repeated Data and Memorization Dynamics}

Closely related is the study of memorization dynamics, particularly how repeated data affects model behavior. \citet{hernandez2022scalinglawsinterpretabilitylearning} trained models where a small portion of the data was repeated many times, observing strong double descent phenomena \citep{advani2020high,belkin2019reconciling,adlam2020understanding,bordelon2020spectrum,schaeffer2024double} and showing that repeating just 0.1\% of tokens 100 times could significantly degrade generalization. Studies tracking exact-sequence memorization have shown that larger models not only memorize more content and at a faster rate but also forget less over the course of training \citep{tirumala2022memorization}. \citet{carlini2023quantifying} quantified log-linear relationships between verbatim generation and model size, data duplication count, and prompt length. Other work has explored the feasibility of *forecasting* whether a model will memorize a specific string, finding that accurate prediction is possible but may require a substantial portion of the target model's pretraining compute \citep{biderman2023emergent}. Beyond explicit repetition, \citet{duan2025uncovering} discovered *latent memorization*, where memorized sequences that are not obvious at a final checkpoint can persist and be revealed later, posing privacy risks. Finally, memorization appears to be task-dependent: \citet{wang2025generalization} observed stronger memorization for knowledge-intensive QA, whereas machine translation and mathematical reasoning demonstrated greater novelty. Memorization also interacts with logical reasoning; using dynamically generated puzzles, \citet{xie2025memorizationlargelanguagemodels} showed that models could be fine-tuned to perfectly memorize training examples yet failed on slight variations, even as their genuine reasoning abilities also improved, revealing a complex balance between the two.

\paragraph{Detecting and Proving Contamination}
Another significant area of research focuses on detecting or proving test set contamination in existing models. \citet{oren2023proving} and \citet{ni2025trainingbenchmarkneed} proposed statistical tests with provable control over false positives by testing if a benchmark's canonical ordering is statistically privileged over random shuffles. \citet{shi2024detecting} introduced Min-$k\%$-Prob to determine if a sequence likely appeared in pretraining using only black-box probabilities. Two related works from \citet{golchin2023data,golchin2024time} frame detection as a multiple-choice "quiz" and use temporal information about model training windows versus benchmark release dates, a strategy also used by \citet{roberts2024to}. Broader audits have aimed to quantify leakage and decontamination across a wide range of tasks and models \citep{xu2024benchmarkingbenchmarkleakagelarge,deng2024investigating,li2024opensource}, while \citet{yang2023rethinking} showed that rephrasing benchmark questions can often bypass n-gram filters. In the domain of code generation, \citet{riddell2024quantifying} quantified contamination in popular coding benchmarks and connected the degree of overlap to performance differences. \citet{matton2024leakage} cataloged various channels for leakage and released a dataset (LBPP) to help mitigate these issues. Complementing these audits, \citet{yang2025rethinkingeffectsdatacontamination} systematically tested fine-grained contamination scenarios in code intelligence across different model types, finding that paired contamination substantially affects LLMs under a pretraining-plus-inference paradigm but has limited effect under a pretrain--finetune--inference pipeline. Other work has also provided instruments for detecting the origins of chain-of-thought sequences \citep{li2025diagnosing}.

\paragraph{Preventing Test Set Contamination}
The growing concern over contamination has spurred the development of new methods for creating benchmarks. These include dynamically updated benchmarks \citep{jain2025livecodebench, xia2024top, zhang2025dynamicbenchmarkconstructionevaluating, qian2024varbenchrobustlanguagemodel} and private or restricted-access benchmarks \citep{zhang2024carefulexaminationlargelanguage, glazer2025frontiermathbenchmarkevaluatingadvanced}. Recently, \citet{nie2025uqassessinglanguagemodels} released a benchmark consisting of unsolved scientific questions, which, by its nature, prevents models from being trained on the correct solutions.

\paragraph{Retrieval- and Agent-Time Contamination}
As model evaluation evolves from static prompting to using tool-augmented agents, the risk of contamination expands. \citet{han2025searchtimedatacontamination} introduced search-time contamination, where an agent retrieves benchmark questions and answers from the web during its evaluation process, which can artificially inflate its performance.

\paragraph{Membership Inference Attacks}
The field of Membership Inference Attacks (MIA) aims to determine if a specific data point was used to train a model, given only access to the model itself \citep{shokri2017membership}. This is highly relevant to contamination, as detection can be viewed as an MIA problem. While the MIA literature is extensive in computer vision \citep{yeom2018privacy, salem2018ml, sablayrolles2019white, jagielski2024students}, it has more recently been applied to language models \citep{carlini2021extracting,zarifzadeh2023low,shi2024detecting,mattern2023membership,li2023mope}. However, progress in sequence-level MIA for language models has been complicated by issues such as flawed evaluations \citep{meeus2024inherent,zhang2024membership,jiang2025a}. \citet{duan2024membership} argue that membership can be inherently "blurry" for natural language. \citet{das2024blind} and \cite{meeus2024inherent} report that existing MIA testbeds suffer from distribution shifts. \citet{kong2023can} refute MIAs with a theoretical attack, and \citet{liu2025language} and \citet{mangaokar2025really} demonstrate fundamental limitations and exploits of n-gram based methods. Due to these challenges, recent work explores strengthening the membership signal by using multiple correlated sequences as input \citep{maini2021dataset,kandpal2023user,maini2024llm}, which aligns more closely with detecting contamination of an entire test set rather than a single example \citep{golchin2023data,oren2023proving}.

\clearpage

\section{Pretraining Implementation Details}
\label{app:sec:pretraining_implementation_details}

\paragraph{Model Architecture}
We pretrained Qwen 3 \citep{yang2025qwen3technicalreport} architecture causal language models from random initialization.
Table~\ref{tab:model_architectures} shows the depth (number of layers) and width (hidden size) configurations for each model size.
The intermediate size for the feed-forward layers follows Qwen 3's formula: $256 \cdot \lfloor (255 + \lfloor 8 \cdot \text{hidden\_size} / 3 \rfloor) / 256 \rfloor$.
All models used Flash Attention 2 \citep{dao2023flashattention2fasterattentionbetter} and bfloat16 precision.

\begin{table}[h]
\centering
\caption{Model architecture configurations following Qwen 3 scaling patterns.}
\label{tab:model_architectures}
\begin{tabular}{lcc}
\toprule
Parameters & Num. Layers & Hidden Size \\
\midrule
34M  & 3  & 96  \\
62M  & 5  & 160 \\
93M  & 6  & 224 \\
153M & 9  & 320 \\
344M & 14 & 576 \\
\bottomrule
\end{tabular}
\end{table}

\paragraph{Optimizer and Learning Rate}
We used the AdamW optimizer \citep{loshchilovde2019adamw} with HuggingFace defaults (i.e., $\beta_1 = 0.9$, $\beta_2 = 0.999$, $\epsilon = 10^{-8}$, weight decay $= 0$).
We used linear warmup for 250 steps followed by cosine annealing to zero.
Following \citet{shuai2024scalinglawlanguagemodels}, we scaled the batch size with the total number of training tokens $D$ as:
\begin{equation}
    \text{tokens per optimizer step} = 3.24 \times 10^3 \times D^{0.264}.
\end{equation}
The learning rate was scaled with the square root of the batch size: $\eta = 10^{-6} \times \sqrt{\text{tokens per optimizer step}}$.
Gradients were clipped to a maximum norm of 1.0.

\paragraph{Data Mixing and Contamination}
For each configuration, we created a pretraining corpus by mixing documents from FineWeb-Edu-Dedup \citep{penedo2024finewebdatasetsdecantingweb} with replicated copies of the MATH test set.
The MATH test set was formatted using the template from EleutherAI's LM Evaluation Harness: ``Problem: \{problem\}\textbackslash n\textbackslash nSolution: \{solution\}''.
For a given number of test set replicas $R$, we:
(1) replicated the tokenized MATH test set $R$ times,
(2) computed the remaining token budget as $D_{\text{corpus}} = D_{\text{total}} - R \times |\text{MATH test set}|$,
(3) sampled documents from FineWeb-Edu-Dedup to fill $D_{\text{corpus}}$ tokens, and
(4) shuffled the combined dataset.
This ensures that total training tokens remain constant across contamination levels, isolating the effect of contamination from the effect of additional data.
Each sequence was truncated to a maximum length of 2048 tokens and terminated with an EOS token.

\paragraph{Distributed Training}
Training used PyTorch's DistributedDataParallel (DDP) with the NCCL backend.
All experiments were logged to Weights \& Biases, and trained models were uploaded to HuggingFace Hub.

Our code is available in our \href{https://github.com/RylanSchaeffer/KoyejoLab-Test-Set-Contam-Generative-Evals}{public GitHub repository}.

\paragraph{Existing Asset Licenses}
We cite or name the existing assets used in the study. The MATH benchmark is distributed from the authors' repository under the \href{https://github.com/hendrycks/math/blob/main/LICENSE}{MIT License}. The FineWeb-Edu-Dedup subset we use through the SmolLM Corpus is tagged \href{https://huggingface.co/datasets/HuggingFaceTB/smollm-corpus/tree/main/fineweb-edu-dedup}{ODC-By}; FineWeb-Edu is likewise released under ODC-By and is subject to Common Crawl terms. Public Qwen 3 model materials are distributed under \href{https://huggingface.co/Qwen/Qwen3-0.6B/blob/main/LICENSE}{Apache-2.0}; our experiments use the Qwen 3 architecture from random initialization rather than pretrained Qwen weights. EleutherAI's LM Evaluation Harness is distributed under the \href{https://github.com/EleutherAI/lm-evaluation-harness/blob/main/LICENSE.md}{MIT License}, Flash Attention 2 under \href{https://github.com/Dao-AILab/flash-attention/blob/main/LICENSE}{BSD-3-Clause}, and PyTorch under a \href{https://github.com/pytorch/pytorch/blob/main/LICENSE}{BSD-style license}. Model hosting on HuggingFace Hub is subject to the \href{https://huggingface.co/terms-of-service}{HuggingFace Terms of Service}.

\clearpage

\section{Math Verify Scoring Bug}
\label{app:sec:math_verify_bug}

During our experiments, we discovered that EleutherAI's Language Model Evaluation Harness \citep{gao2024evalharness} contained a bug in its Math Verify scoring implementation that caused systematically incorrect results.
This bug affected the \texttt{minerva\_math} task and related mathematics evaluation tasks.

\paragraph{The Problem}
The \texttt{minerva\_math} task's utility code called \texttt{remove\_boxed()} on extracted answers \emph{before} passing them to \texttt{math\_verify.parse()}.
The original implementation was:
\begin{verbatim}
res = verify(parse(doc["answer"]), parse(candidates))
\end{verbatim}
where \texttt{doc["answer"]} had already been processed to remove the \LaTeX{} \verb|\boxed{}| wrapper.

The \texttt{math\_verify.parse()} function relies on the \verb|\boxed{}| notation to properly identify and extract mathematical expressions.
When the boxed notation is stripped beforehand:
\begin{itemize}
    \item \texttt{parse("\textbackslash dfrac\{9\}\{7\}")} returns empty results (parsing fails)
    \item \texttt{parse("\textbackslash boxed\{\textbackslash dfrac\{9\}\{7\}\}")} correctly returns the parsed expression
\end{itemize}

This caused Math Verify scores to return 0 even when the model's answer was mathematically correct.
As a diagnostic, we evaluated the benchmark's own gold reference solutions and observed Math Verify scores of approximately 70\%---a clear indication that the scoring mechanism itself was flawed rather than the solutions.

\paragraph{The Fix}
\href{https://github.com/EleutherAI/lm-evaluation-harness/issues/3210}{We reported this issue} and \href{https://github.com/EleutherAI/lm-evaluation-harness/pull/3259}{a fix was merged} that passes the full solution text to \texttt{math\_verify.parse()}, allowing it to use its built-in extraction heuristics:
\begin{verbatim}
res = verify(gold=parse(doc["solution"]), target=parse(candidates))
\end{verbatim}

By using \texttt{doc["solution"]} (the complete solution with \verb|\boxed{}| intact) instead of the pre-processed \texttt{doc["answer"]}, the parser can correctly identify and extract the mathematical expressions.

\paragraph{Implications}
This bug was present in versions of the evaluation harness prior to v0.4.8 (August 2025).
Any research reporting MATH benchmark scores using Math Verify from affected versions may have systematically underestimated model performance.
We recommend that researchers using these benchmarks verify they are using a corrected version of the evaluation code.

\clearpage

\section{Exceptions to In-Context / Sequence Scaling Laws}
\label{app:sec:exceptions_to_sequence_scaling_laws}

In Sec.~\ref{sec:inference}, we reported that most model sizes and number of test set replicas exhibit in-context scaling laws (Eq.~\ref{eqn:in_context_scaling_law}).
While the power law plus irreducible error model provides an excellent fit for the majority of experimental conditions, we observe a systematic deviation in specific edge cases.
In these instances, the negative log-likelihood (NLL) \textit{increases} with token index $t$ rather than decaying.
We identified significant positive slopes in the NLL-vs-token trajectory for two distinct groups of models.

\paragraph{Group I: Capacity Limitations (Uncontaminated Regime)}
For the smallest uncontaminated model (34M, $R=0$), the NLL increases monotonically from token 1 to token 645. This behavior is consistent with limited effective context windows in small architectures. Without the aid of memorization ($R=0$), the 34M parameter model struggles to model long-range dependencies, causing prediction quality to degrade as the necessary context grows beyond its effective capacity. Larger uncontaminated models (62M+) do not exhibit this behavior, showing standard NLL decay.

\paragraph{Group II: Selection Bias (High Contamination Regime)}
The most pronounced uptick occurs in the largest, most contaminated models (e.g., 344M, $R=3162$), where NLL is extremely low ($\sim 10^{-3}$) but rises sharply at late token indices. Our analysis suggests this is driven by \textbf{selection bias inherent to the dataset structure}.

Because the MATH dataset contains solutions of varying lengths, the set of problems contributing to the NLL at $t=800$ is a small, non-random subset of the problems at $t=100$.
\begin{itemize}
    \item \textbf{Survivor Bias:} Only the 100 longest solutions in the test set reach token 800.
    \item \textbf{Differential Memorizability:} We observe a strong correlation between solution length and residual NLL in the high-contamination regime. The uptick indicates that these long surviving sequences are systematically ``harder'' to memorize than the shorter sequences that drop out earlier.
\end{itemize}

We verified that these deviations are not artifacts of data corruption. The token-level sample counts are consistent across all runs, with $100\%$ of sequences (N=5,000) present at token 0, dropping to $2\%$ (N=100) at token 800. This attrition is strictly deterministic, governed by the length distribution of the MATH dataset.

\begin{figure*}[h!]
    \centering
    \includegraphics[width=\linewidth]{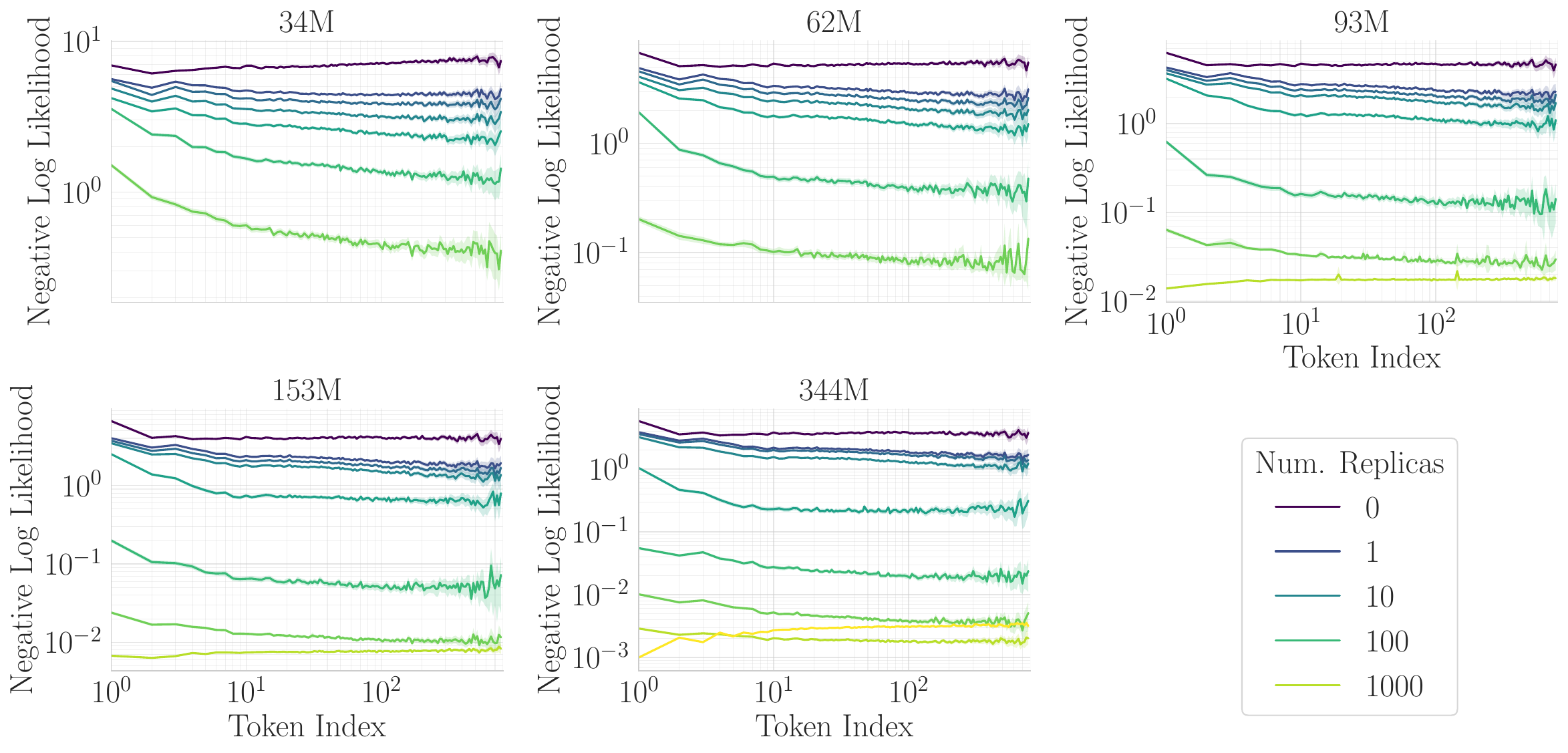}
    \caption{\textbf{Deviations from In-Context / Sequence Scaling Laws.} Two groups of models exhibit increasing negative log likelihoods with increasing token index: smaller uncontaminated models and larger massively contaminated models.}
\end{figure*}

\clearpage

\begin{figure*}[t!]
    \centering
    \includegraphics[width=\linewidth]{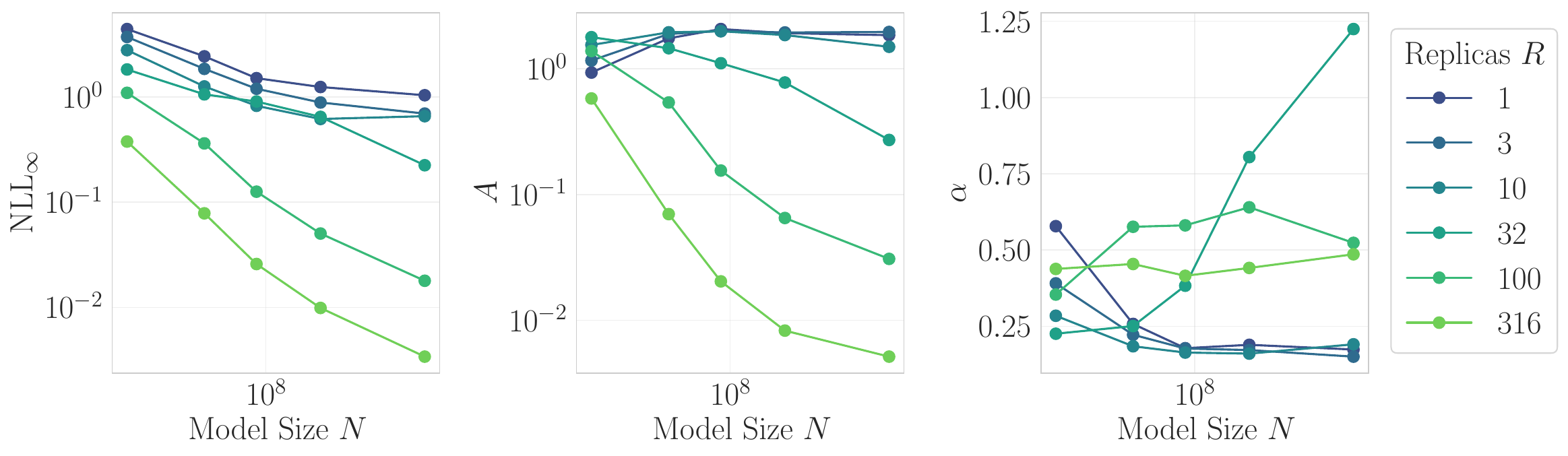}
    \caption{\textbf{Fit Parameters for In-Context / Sequential Scaling Laws by Model Size and Number of Test Set Replicas.} }
\end{figure*}

\clearpage

\begin{figure*}[t!]
    \centering
    \includegraphics[width=\linewidth]{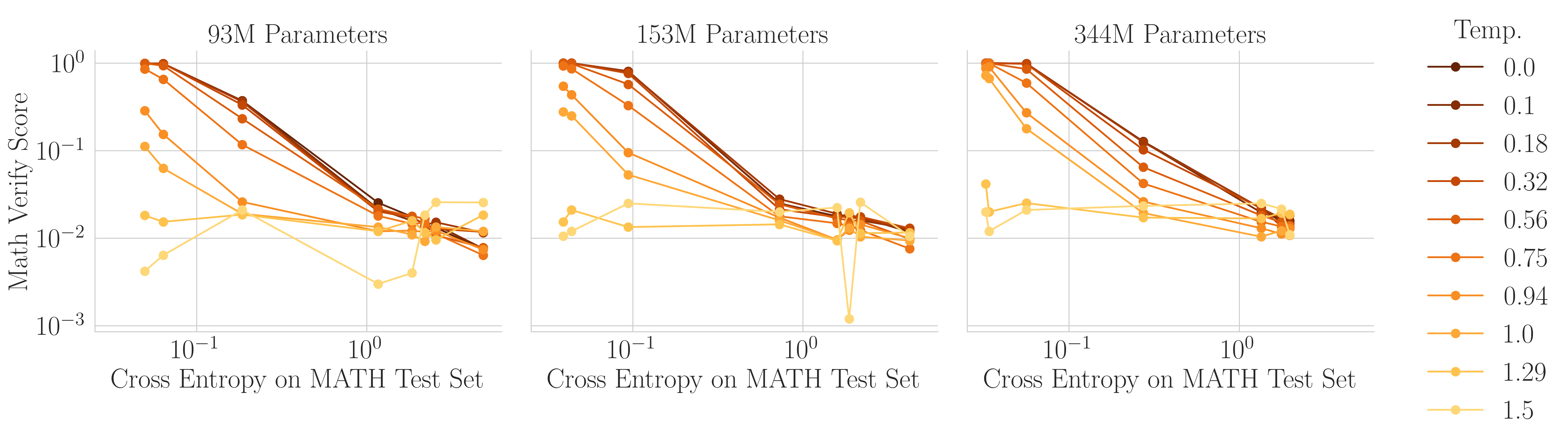}
    \caption{\textbf{Math Verify Score Is Correlated with Pretraining Loss.} Math Verify scores correlate strongly with the cross entropy loss achieved on the MATH Test Set during training, where differences in these graphs are attributable to increased repetition on the benchmark test set.  The correlation is significantly weaker for high temperatures and falls to nearly $0$ for temperatures above $1.0$.  }
    \label{fig:loss_and_math_verify}
\end{figure*}

\clearpage

\section{Discussion (Continued)}
\label{app:sec:discussion}

\paragraph{Additional Limitations}
We focused on a single generative benchmark, MATH, to enable automatic verification and controlled contamination. Consequently, our findings may not fully capture how contamination behaves for other tasks such as coding or creative writing.
Additionally, our experiments utilized decoder-only dense transformer models (Qwen 3) up to 344M parameters; results at this scale may not extrapolate to larger models or other architectures.
Finally, our pretraining corpus represents a specific mixture of web-crawl data; specialized or heavily filtered corpora could alter the base difficulty of memorization and the specific thresholds at which contamination effects become visible.

\paragraph{Future Directions}
Several avenues for future research:
\begin{itemize}
    \item \textbf{Reconciling Memorization Thresholds:} Our discovery that a single test set replica drives loss below the irreducible error contrasts with recent assertions that single-shot verbatim memorization is an ``illusion'' \citep{huang2024demystifying} or that membership inference is fundamentally capped \citep{hayes2025exploring}. Future work should investigate what the explanatory differences are; one possibility is that memorization depends on the amount relative to corpus size \citep{schaeffer2025doseresponse, wei2025hubblemodelsuiteadvance}.
    \item \textbf{Inference-Time Detection Methods:} Since high-temperature sampling and long-context generation disproportionately harm contaminated models, future work could develop lightweight ``stress tests'' that sweep temperature or solution length to detect contamination without access to the pretraining corpus.
    \item \textbf{Architectural Susceptibility:} We studied dense models, but it remains an open question whether Mixture-of-Experts (MoEs) or State Space Models (SSMs) exhibit different memorization patterns.
    \item \textbf{Harder Benchmarks:} We hypothesize that contamination efficacy is inversely correlated with problem compressibility. Investigating whether harder benchmarks (e.g., AIME) require higher ``doses'' of leakage to achieve the same performance boost would be a valuable contribution to benchmark design.
\end{itemize}

\end{document}